\documentclass[10pt,twocolumn]{article}

\usepackage[T1]{fontenc}
\usepackage[utf8]{inputenc}
\usepackage{amsmath,amssymb}
\usepackage{booktabs}
\usepackage{array}
\usepackage{makecell}
\usepackage[table]{xcolor}
\usepackage{hyperref}
\usepackage[numbers,sort&compress]{natbib}
\usepackage{enumitem}
\usepackage{microtype}
\usepackage[section]{placeins}
\usepackage{float}
\usepackage{balance}
\usepackage{algorithm}
\usepackage{algpseudocode}
\usepackage{tikz}
\usepackage{pgfplots}
\usetikzlibrary{arrows.meta,positioning,shapes.geometric}
\pgfplotsset{compat=1.18}
\usepackage[margin=0.75in,top=0.85in,bottom=0.9in]{geometry}

\definecolor{linkpink}{RGB}{220, 20, 120}
\definecolor{deepblue}{RGB}{0,51,102}
\definecolor{bestblue}{RGB}{222,235,247}
\definecolor{targetgreen}{RGB}{46,125,75}
\definecolor{warningorange}{RGB}{202,111,30}
\definecolor{softgray}{RGB}{246,247,249}
\newcommand{\best}[1]{%
  \cellcolor{bestblue}\begingroup\bfseries\boldmath #1\endgroup%
}

\newcommand{\keypoint}[1]{%
  \begin{center}
  \fcolorbox{deepblue}{softgray}{%
    \parbox{0.91\columnwidth}{\small\centering\textbf{Design principle.} #1}%
  }
  \end{center}
}

\hypersetup{
  colorlinks=true,
  linkcolor=deepblue,
  citecolor=deepblue,
  urlcolor=deepblue,
  pdfauthor={Jupiter Yao},
  pdftitle={H2DF: Calibrated Target Bands for Controllable Language-Model Unlearning}
}
\setlist{nosep,leftmargin=*}
\newcommand{\Lcal}{\mathcal{L}}
\newcommand{\Fcal}{\mathcal{F}}
\newcommand{\method}{\textsc{H2DF}}
\newcolumntype{L}[1]{>{\raggedright\arraybackslash}p{#1}}

\title{\vspace{-1.2em}\textbf{H2DF: Calibrated Target Bands for\\
Controllable Language-Model Unlearning}}
\author{
Jupiter Yao\\
Cornell University\\
\texttt{jy2228@cornell.edu}
}
\date{}

\begin{document}
\maketitle

\begin{abstract}
Gradient ascent (GA) is a simple language-model unlearning baseline, but it
continues increasing forget loss after useful forgetting has occurred. This
unbounded pressure can damage retained utility and make forgotten examples
abnormally high-loss. We introduce \method, a calibrated target-band objective
that applies GA-like pressure below a data-dependent lower target and suppresses,
or reverses, that pressure after the target region is reached. The targets are
estimated from matched non-member examples before training, so the base objective
requires one forget-batch forward/backward pass per optimization step; our
reported TOFU variant adds sparse retain replay every tenth step. We instantiate
a one-sided objective for TOFU question answering and a two-sided objective for
MUSE-News language modeling. Across three TOFU seeds, \method{} obtains a forget
NLL of $1.528\pm0.049$ and retain NLL of $1.165\pm0.047$, matching GradDiff
while using retain replay only every tenth step and showing substantially lower
variability than GA. On MUSE-News, \method{} improves retention over GA and NPO
and is competitive with SimNPO, but does not dominate it. Ablations show that the
upper target reduces overshoot and that validation-based checkpoint selection is
important. These results support calibrated target bands as a practical mechanism
for controlling forgetting strength, while also exposing sensitivity to random
seed and calibration quality. \href{https://github.com/Shdakudow/H2DF-Calibrated-Target-Bands-for-Controllable-Language-Model-Unlearning}
{\textcolor{linkpink}{\textbf{Source Code and Models}}}
\end{abstract}

\begin{figure}[t]
\centering
\begin{tikzpicture}
\begin{axis}[
  width=\columnwidth,
  height=0.52\columnwidth,
  xmin=0, xmax=3.6,
  ymin=-0.65, ymax=1.15,
  xlabel={Forget score $s_i(\theta)$},
  ylabel={Pressure to increase $s_i$},
  xtick={1.4,2.2},
  xticklabels={$a_i$,$b_i$},
  ytick={-0.5,0,0.5,1},
  axis lines=left,
  grid=major,
  grid style={draw=gray!18},
  tick label style={font=\scriptsize},
  label style={font=\scriptsize},
  legend style={
    font=\tiny,
    at={(0.5,-0.30)},
    anchor=north,
    legend columns=3,
    draw=none,
    fill=none,
    column sep=3pt
  },
  samples=120
]
\path[fill=targetgreen!13] (axis cs:1.4,-0.65)
  rectangle (axis cs:2.2,1.15);
\addplot[warningorange,very thick,domain=0:3.6] {1};
\addlegendentry{GA}
\addplot[gray!70!black,thick,dashed,domain=0:3.6]
  {1/(1+exp(2.5*(x-1.45)))};
\addlegendentry{Preference}
\addplot[deepblue,very thick,domain=0:3.6]
  {1/(1+exp((x-1.4)/0.25))-0.5/(1+exp((2.2-x)/0.25))};
\addlegendentry{\method}
\draw[targetgreen,densely dotted,thick]
  (axis cs:1.4,-0.65) -- (axis cs:1.4,1.15);
\draw[targetgreen,densely dotted,thick]
  (axis cs:2.2,-0.65) -- (axis cs:2.2,1.15);
\node[font=\tiny,text=targetgreen!70!black,anchor=south]
  at (axis cs:1.8,-0.59) {target band};
\end{axis}
\end{tikzpicture}
\caption{Conceptual forget pressure. GA remains persistent; preference-style
objectives attenuate pressure implicitly
\citep{zhang2024negative,fan2024simplicity}; \method{} pushes below $a_i$,
saturates in the calibrated band, and can restore above $b_i$. Curves use
illustrative parameters.}
\label{fig:first-page-contrast}
\end{figure}
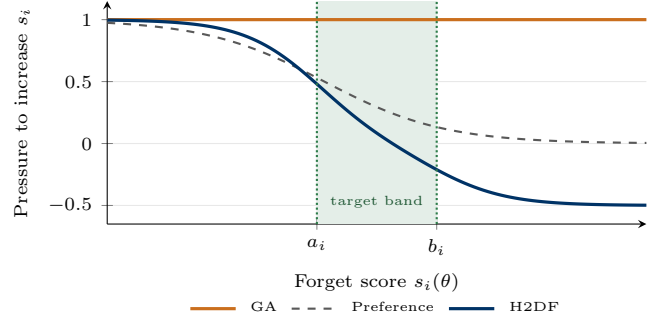

\section{Introduction}

Large language models may memorize private, copyrighted, or otherwise undesirable
training data. Machine unlearning seeks to remove the influence of selected data
without retraining from scratch \citep{cao2015towards}. For large models, a
common baseline is gradient ascent on the forget set \citep{yao2023large}. Given
a forget batch $B_f$ and per-example negative log-likelihood (NLL)
$\ell_i(\theta)$, GA minimizes
\begin{equation}
  \Lcal_{\mathrm{GA}}(\theta)
  =-\frac{1}{|B_f|}\sum_{i\in B_f}\ell_i(\theta).
  \label{eq:ga}
\end{equation}
GA is inexpensive, but it treats forgetting as an infinite direction: the
objective continues to increase NLL even after the model no longer reproduces the
targeted content. This can cause catastrophic utility loss and abnormal high-loss
artifacts. Preference-based methods such as NPO and SimNPO temper this behavior
\citep{zhang2024negative,fan2024simplicity}, while retain-aware methods such as
GradDiff add utility-preserving supervision at additional online cost.
Figure~\ref{fig:first-page-contrast} summarizes the central distinction.

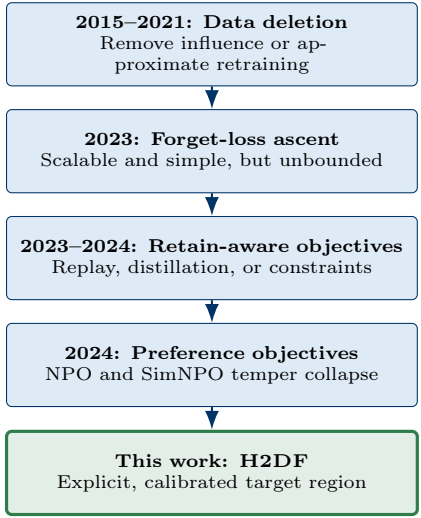
\begin{figure}[t]
\centering
\begin{tikzpicture}[
  stage/.style={
    draw=deepblue,
    rounded corners=2pt,
    minimum height=11mm,
    text width=52mm,
    align=center,
    font=\scriptsize
  },
  main/.style={stage,fill=bestblue},
  ours/.style={stage,fill=targetgreen!13,draw=targetgreen,very thick},
  flow/.style={-{Latex[length=2.5mm]},thick,draw=deepblue}
]
\node[main] (delete) at (0,0)
  {\textbf{2015--2021: Data deletion}\\Remove influence or approximate retraining};
\node[main,below=3mm of delete] (ga)
  {\textbf{2023: Forget-loss ascent}\\Scalable and simple, but unbounded};
\node[main,below=3mm of ga] (retain)
  {\textbf{2023--2024: Retain-aware objectives}\\Replay, distillation, or constraints};
\node[main,below=3mm of retain] (pref)
  {\textbf{2024: Preference objectives}\\NPO and SimNPO temper collapse};
\node[ours,below=3mm of pref] (h2df)
  {\textbf{This work: \method}\\Explicit, calibrated target region};
\draw[flow] (delete) -- (ga);
\draw[flow] (ga) -- (retain);
\draw[flow] (retain) -- (pref);
\draw[flow] (pref) -- (h2df);
\end{tikzpicture}
\caption{Selective progression of unlearning objectives. \method{} makes the
stopping region explicit rather than pushing indefinitely away from the
original likelihood.}
\label{fig:history}
\end{figure}

Figure~\ref{fig:history} gives a selective view of the objective-design path
that motivates our method. Classical work focused on deleting training influence
or approximating retraining \citep{cao2015towards,ginart2019making,
bourtoule2021machine}. LLM unlearning made direct forget-loss maximization a
scalable baseline \citep{yao2023large}; retain-aware and preference objectives
then sought to limit utility collapse \citep{chen2023unlearn,
zhang2024negative,fan2024simplicity}. \method{} continues this progression by
making the stopping region explicit and calibrating it from non-members.

We study a direct alternative: forgetting should be a calibrated region rather
than an unbounded direction. \method{} estimates a target loss interval from
matched non-member data and optimizes each forget example toward that interval.
Below the interval, its gradient follows GA; inside the interval, pressure
saturates; above the interval, an optional upper penalty discourages overshoot.
Calibration is performed once, and training uses only the current model's
forget-batch losses.

\keypoint{\emph{``Forgetting should be a calibrated region, not an infinite
direction.''}}

Our contributions are:
\begin{enumerate}[label=(\roman*)]
  \item a general target-band objective that recovers GA-like updates when
        forgetting is insufficient and suppresses them after a calibrated target
        is reached;
  \item task-specific instantiations for TOFU \citep{maini2024tofu} and MUSE
        \citep{shi2024muse};
  \item experiments against GA, GradDiff, NPO, and SimNPO, including three-seed
        TOFU results, MUSE-News official metrics, and target-band ablations; and
  \item an empirical analysis of overshoot, checkpoint selection, and seed
        sensitivity in short-budget LoRA unlearning.
\end{enumerate}

\section{H2DF Method}
\label{sec:method}

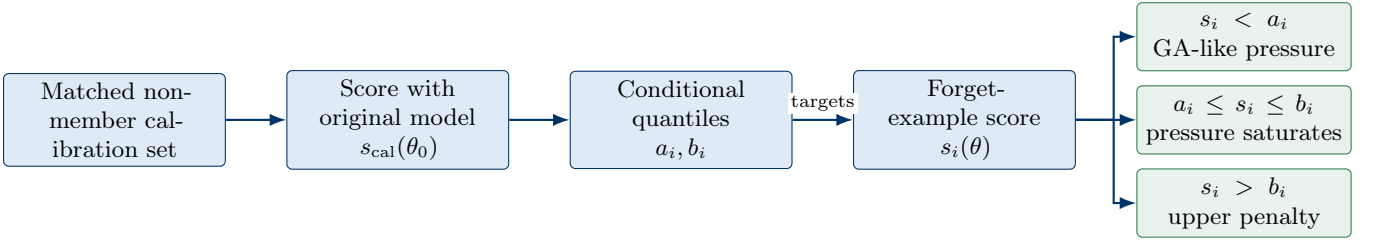
\begin{figure*}[t]
\centering
\begin{tikzpicture}[
  node distance=7mm and 8mm,
  box/.style={
    draw=deepblue,
    rounded corners=2pt,
    fill=bestblue,
    minimum height=10mm,
    text width=27mm,
    align=center,
    font=\small
  },
  action/.style={
    draw=targetgreen,
    rounded corners=2pt,
    fill=targetgreen!10,
    minimum height=9mm,
    text width=26mm,
    align=center,
    font=\small
  },
  arrow/.style={-{Latex[length=2.2mm]},thick,draw=deepblue}
]
\node[box] (cal) {Matched non-member calibration set};
\node[box, right=of cal] (score) {Score with original model\\$s_{\mathrm{cal}}(\theta_0)$};
\node[box, right=of score] (quant) {Conditional quantiles\\$a_i,b_i$};
\node[box, right=of quant] (train) {Forget-example score\\$s_i(\theta)$};
\node[action, right=of train, yshift=11mm] (low) {$s_i<a_i$\\GA-like pressure};
\node[action, right=of train] (inside) {$a_i\leq s_i\leq b_i$\\pressure saturates};
\node[action, right=of train, yshift=-11mm] (high) {$s_i>b_i$\\upper penalty};
\draw[arrow] (cal) -- (score);
\draw[arrow] (score) -- (quant);
\draw[arrow] (quant) -- node[midway, above=2pt, font=\scriptsize, fill=white, inner sep=1pt]
{targets} 
(train);
\draw[arrow] (train.east) -- ++(5mm,0) |- (low.west);
\draw[arrow] (train.east) -- (inside.west);
\draw[arrow] (train.east) -- ++(5mm,0) |- (high.west);
\end{tikzpicture}
\caption{\method{} workflow. Calibration converts matched non-member scores into
per-example target regions before training. During unlearning, examples below
the lower target receive GA-like pressure, examples inside the target region
receive little pressure, and the optional upper term discourages overshoot.}
\label{fig:overview}
\end{figure*}

\subsection{General Objective}

Let $s_i(\theta)$ be a score for forget example $i$, with larger values denoting
stronger forgetting. In our experiments, $s_i$ is token-averaged NLL. Let
$[a_i,b_i]$ be a target interval estimated from a held-out non-member calibration
set, conditional on observable features $c_i$ such as sequence length, domain,
and original-model loss. We define
\begin{equation}
  a_i=Q_{\rho_{\mathrm{low}}}(s_{\mathrm{cal}}\mid c_i),\qquad
  b_i=Q_{\rho_{\mathrm{high}}}(s_{\mathrm{cal}}\mid c_i),
  \label{eq:targets}
\end{equation}
where $Q_\rho$ is an empirical quantile.
Figure~\ref{fig:overview} summarizes how these fixed targets alter the online
forget update.

Using the smooth hinge $h_\tau(u)=\tau\log(1+\exp(u/\tau))$, the objective is
\begin{equation}
\begin{aligned}
  \Lcal_{\mathrm{H2DF}}(\theta)
  ={}&\frac{1}{|B_f|}\sum_{i\in B_f}
  \left[h_\tau(a_i-s_i(\theta))
  +\gamma h_\tau(s_i(\theta)-b_i)\right]\\
  &+\lambda\|\theta_{\mathrm{LoRA}}\|_2^2 .
\end{aligned}
\label{eq:h2df}
\end{equation}
The lower term penalizes insufficient forgetting. The upper term, weighted by
$\gamma\geq0$, penalizes excessive loss; $\gamma=0$ gives a one-sided objective.
The targets are precomputed, so Eq.~\eqref{eq:h2df} has the same online
forward/backward structure as GA.

\subsection{Gradient Behavior}

For the one-sided objective,
\begin{equation}
\begin{aligned}
  \nabla_\theta h_\tau(a_i-s_i(\theta))
  ={}& -\alpha_i(\theta)\nabla_\theta s_i(\theta), \\
  \alpha_i(\theta)
  ={}& \sigma\!\left(
    \frac{a_i-s_i(\theta)}{\tau}
  \right).
\end{aligned}
\label{eq:gradient}
\end{equation}
Thus $\alpha_i\approx1$ when $s_i\ll a_i$, recovering the direction of GA, and
$\alpha_i\to0$ after $s_i$ exceeds the lower target. With an upper target,
\begin{equation}
  \frac{\partial\Lcal_i}{\partial s_i}
  =-\sigma\!\left(\frac{a_i-s_i}{\tau}\right)
  +\gamma\sigma\!\left(\frac{s_i-b_i}{\tau}\right),
\end{equation}
so sufficiently large losses receive a restoring gradient. This bounded
coefficient does not bound the model Jacobian $\nabla_\theta s_i$; the claim is
specifically that the objective removes GA's persistent scalar incentive to
increase loss.

\subsection{Task Instantiations}

\paragraph{TOFU.}
For question $q_i$ and answer tokens $y_{i,1:T_i}$, the score is answer-only NLL:
\begin{equation}
  s_i(\theta)=-\frac{1}{T_i}\sum_{t=1}^{T_i}
  \log p_\theta(y_{i,t}\mid q_i,y_{i,<t}).
\end{equation}
We use a one-sided target
\begin{equation}
  a_i=\max\!\left\{
  s_i(\theta_0)+\Delta_{\min},
  Q_\rho(s_{\mathrm{cal}}(\theta_0)\mid c_i)
  \right\},
  \label{eq:tofu-target}
\end{equation}
which combines a minimum displacement from the original model with a non-member
quantile.

\paragraph{MUSE-News.}
For text $x_{i,1:T_i}$, $s_i$ is token-averaged language-modeling NLL. We use
the two-sided targets in Eq.~\eqref{eq:targets}. The goal is not to maximize
perplexity, but to move forgotten text toward the loss regime of matched
non-members.

\subsection{Illustrative Unlearning Example}

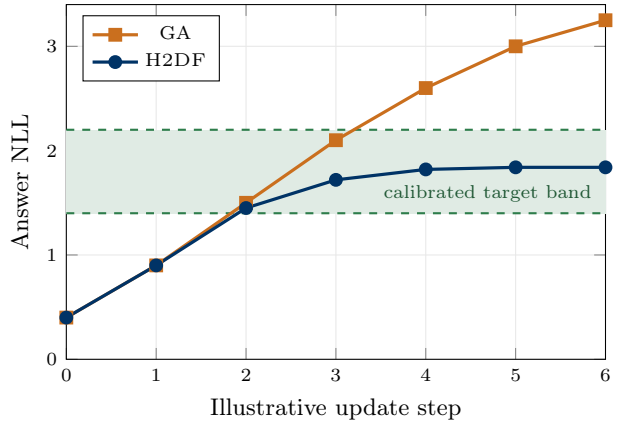
\begin{figure}[t]
\centering
\begin{tikzpicture}
\begin{axis}[
  width=\columnwidth,
  height=0.72\columnwidth,
  xmin=0, xmax=6,
  ymin=0, ymax=3.4,
  xlabel={Illustrative update step},
  ylabel={Answer NLL},
  xtick={0,1,2,3,4,5,6},
  grid=major,
  grid style={draw=gray!18},
  tick label style={font=\scriptsize},
  label style={font=\small},
  legend style={
    font=\scriptsize,
    at={(0.03,0.97)},
    anchor=north west
  }
]
\path[fill=targetgreen!15] (axis cs:0,1.4) rectangle (axis cs:6,2.2);
\draw[targetgreen,dashed,thick] (axis cs:0,1.4) -- (axis cs:6,1.4);
\draw[targetgreen,dashed,thick] (axis cs:0,2.2) -- (axis cs:6,2.2);
\addplot[warningorange,very thick,mark=square*] coordinates {
  (0,.4) (1,.9) (2,1.5) (3,2.1) (4,2.6) (5,3.0) (6,3.25)
};
\addlegendentry{GA}
\addplot[deepblue,very thick,mark=*] coordinates {
  (0,.4) (1,.9) (2,1.45) (3,1.72) (4,1.82) (5,1.84) (6,1.84)
};
\addlegendentry{\method}
\node[font=\scriptsize,anchor=south east,text=targetgreen!70!black]
  at (axis cs:5.95,1.42) {calibrated target band};
\end{axis}
\end{tikzpicture}
\caption{Illustrative per-example dynamics. For a synthetic memorized answer
with target band $[1.4,2.2]$, GA continues increasing NLL, whereas \method{}
approaches the non-member-like region and saturates. Values are pedagogical,
not measured experiment outputs.}
\label{fig:example}
\end{figure}

Figure~\ref{fig:example} illustrates the mechanism on a synthetic TOFU-style
question. Suppose the original model assigns answer NLL $0.4$ to a memorized
answer, while matched non-member answers define a target band $[1.4,2.2]$.
Both GA and \method{} initially increase the answer NLL. After \method{} enters
the band, its lower-target pressure vanishes and the upper term prevents large
overshoot. GA has no corresponding stopping signal and continues increasing
loss. The numerical trajectory is illustrative rather than an experimental
trace; it visualizes the gradient behavior in Eq.~\eqref{eq:gradient}.

\FloatBarrier

\section{Experimental Setup}
\label{sec:setup}

\paragraph{TOFU.}
We use the TOFU forget-5\% split and the remaining authors for retention and
calibration. The base is \texttt{Qwen2.5-1.5B-Instruct} with a supervised TOFU
adapter, followed by rank-8 LoRA unlearning. Runs use AdamW, learning rate
$10^{-4}$, batch size 4, 500 steps, cosine scheduling, $\tau=0.25$, $\rho=0.7$,
$\Delta_{\min}=0.5$, and $\lambda=10^{-4}$. We report seeds $42$, $43$, and $44$
where available. Evaluation covers 200 forget and 3,420 retain examples; exact
match is close to zero for all methods, so token F1 is more discriminative. The
reported \method{} variant applies a retain batch every 10 steps; GradDiff
applies retain supervision every step.

\paragraph{MUSE-News.}
For MUSE-News, we use \nolinkurl{muse-bench/MUSE-News_target} as the target
model/checkpoint and the \nolinkurl{NousResearch/Llama-2-7b-hf} tokenizer. Rank-8
LoRA runs use AdamW, learning rate $2\times10^{-5}$, effective batch size 4, 50
steps, $\tau=0.25$, $\gamma=0.5$,
$(\rho_{\mathrm{low}},\rho_{\mathrm{high}})=(0.5,0.9)$, and $\lambda=10^{-4}$.
Full internal evaluation contains 888 forget, 1,778 retain, and 1,521 calibration
examples.

\paragraph{Baselines and selection.}
We compare with the original model, GA, GradDiff, NPO, and SimNPO. All compared
unlearning methods use the same LoRA parameterization. For MUSE, checkpoints are
screened on a fixed 200-example validation subset and selected by proximity to
50\% of examples below the lower target. The screen is used only for checkpoint
selection; reported seed-42 internal results use the full held-out splits.
Official MUSE metrics are reported for the original model, \method, and SimNPO.
Appendix~\ref{app:algorithm} provides the training algorithm, exact
configurations, evidence provenance, and checkpoint-selection rule.

\paragraph{Metrics.}
For TOFU, higher forget NLL and lower forget token F1 indicate stronger
forgetting; lower retain NLL and higher retain token F1 indicate better utility.
For MUSE internal diagnostics, we report forget and retain NLL, loss-based
membership-inference AUC, and the fraction of forget examples inside the
calibrated band. AUC is diagnostic only and does not replace the official MUSE
suite. Official VerbMem-Forget and KnowMem-Forget are lower-is-better;
KnowMem-Retain is higher-is-better. We reproduce PrivLeak with the official
evaluator and preserve its sign convention.

\section{Results}
\label{sec:results}

\subsection{TOFU}

Table~\ref{tab:tofu42} shows the seed-42 tradeoff. SimNPO forgets most strongly
but also incurs the largest utility loss. \method{} and GradDiff occupy a less
aggressive region: \method{} has slightly higher forget NLL and higher retained
token F1 than GradDiff, while their NLLs are nearly identical. GA and NPO are
also nearly identical in this run.

\begin{figure}[t]
\centering
\begin{tikzpicture}
\begin{axis}[
  width=\columnwidth,
  height=0.72\columnwidth,
  xlabel={Retain NLL $\downarrow$},
  ylabel={Forget NLL $\uparrow$},
  xmin=0.95, xmax=1.53,
  ymin=1.35, ymax=1.84,
  grid=major,
  grid style={draw=gray!20},
  legend style={font=\scriptsize,at={(0.02,0.98)},anchor=north west},
  tick label style={font=\scriptsize},
  label style={font=\small},
  error bars/error bar style={line width=0.7pt},
  error bars/error mark options={rotate=90,mark size=2pt}
]
\addplot+[
  only marks, mark=square*, mark size=2.5pt, color=warningorange,
  error bars/.cd, x dir=both, x explicit, y dir=both, y explicit
] coordinates {(1.300,1.595) +- (0.190,0.196)};
\addlegendentry{GA}
\addplot+[
  only marks, mark=triangle*, mark size=3pt, color=gray!70!black,
  error bars/.cd, x dir=both, x explicit, y dir=both, y explicit
] coordinates {(1.162,1.499) +- (0.051,0.043)};
\addlegendentry{GradDiff}
\addplot+[
  only marks, mark=*, mark size=3pt, color=deepblue,
  error bars/.cd, x dir=both, x explicit, y dir=both, y explicit
] coordinates {(1.165,1.528) +- (0.047,0.049)};
\addlegendentry{\method}
\end{axis}
\end{tikzpicture}
\caption{TOFU forgetting--retention tradeoff over seeds 42--44. Points are
means and error bars are sample standard deviations. The preferred region is
toward the upper left. \method{} and GradDiff are stable and closely matched,
whereas GA has substantially larger variance.}
\label{fig:tofu-tradeoff}
\end{figure}
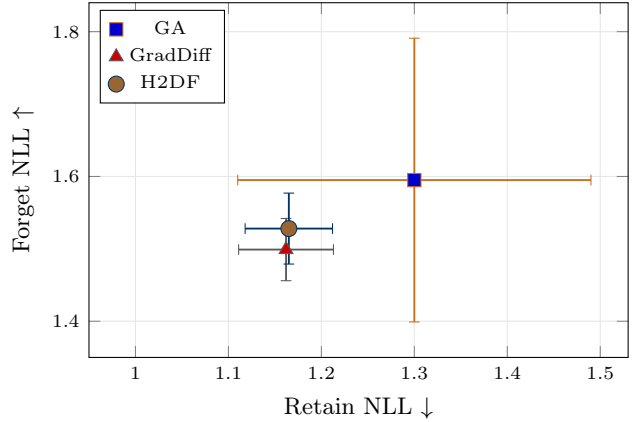

Across three seeds (Table~\ref{tab:tofu3} and Fig.~\ref{fig:tofu-tradeoff}),
\method{} and GradDiff remain close. \method{} achieves somewhat stronger NLL
forgetting and slightly higher retain F1, whereas GradDiff has lower forget F1
and marginally lower retain NLL. GA is more aggressive on average but
substantially more variable; its seed-44 run raises both forget and retain NLL.
The result supports competitive forgetting--retention control, not uniform
dominance.

\begin{table}[H]
\centering
\caption{TOFU seed-42 results. Arrows indicate the preferred direction; best
unlearning values are shaded.}
\label{tab:tofu42}
\small
\renewcommand{\arraystretch}{1.08}
\setlength{\tabcolsep}{5pt}
\begin{tabular}{@{}lcccc@{}}
\toprule
& \multicolumn{2}{c}{Forget} & \multicolumn{2}{c}{Retain} \\
\cmidrule(lr){2-3}\cmidrule(l){4-5}
Method & NLL$\uparrow$ & F1$\downarrow$ & NLL$\downarrow$ & F1$\uparrow$\\
\midrule
Original/SFT & 1.004 & 0.441 & 0.997 & 0.446\\
GA           & 1.560 & 0.312 & 1.280 & 0.334\\
NPO          & 1.561 & 0.310 & 1.282 & 0.333\\
SimNPO       & \best{1.643} & \best{0.287} & 1.365 & 0.309\\
GradDiff     & 1.538 & 0.318 & 1.220 & 0.353\\
\method{}    & 1.565 & 0.327 & \best{1.217} & \best{0.360}\\
\bottomrule
\end{tabular}
\end{table}

\begin{table}[H]
\centering
\caption{TOFU mean $\pm$ sample standard deviation over seeds 42--44.}
\label{tab:tofu3}
\scriptsize
\renewcommand{\arraystretch}{1.10}
\setlength{\tabcolsep}{2.8pt}
\begin{tabular}{@{}lcccc@{}}
\toprule
& \multicolumn{2}{c}{Forget} & \multicolumn{2}{c}{Retain} \\
\cmidrule(lr){2-3}\cmidrule(l){4-5}
Method & NLL$\uparrow$ & F1$\downarrow$ & NLL$\downarrow$ & F1$\uparrow$\\
\midrule
GA & $1.595{\pm}0.196$ & \best{$0.310{\pm}0.042$} & $1.300{\pm}0.190$ & $0.341{\pm}0.054$\\
GradDiff & $1.499{\pm}0.043$ & $0.348{\pm}0.026$ & \best{$1.162{\pm}0.051$} & $0.389{\pm}0.032$\\
\method{} & \best{$1.528{\pm}0.049$} & $0.353{\pm}0.022$ & $1.165{\pm}0.047$ & \best{$0.394{\pm}0.029$}\\
\bottomrule
\end{tabular}
\end{table}

\subsection{MUSE-News}

Table~\ref{tab:muse-internal} reports the full held-out loss diagnostics. GA and
NPO attain the lowest loss-based MIA AUC, but at the cost of the largest retain
degradation. \method{} reduces retain NLL by 0.35 relative to GA and NPO while
maintaining a lower AUC than GradDiff. SimNPO is the closest comparator: it has
better retain NLL and a higher in-band rate, while \method{} has slightly lower
AUC and higher forget NLL. These diagnostics therefore show a competitive, not
dominant, tradeoff.

Official metrics in Table~\ref{tab:muse-official} show that \method{} and SimNPO
are close on knowledge forgetting. SimNPO has stronger verbatim forgetting,
while \method{} retains slightly more knowledge utility. We report PrivLeak
using the official evaluator's sign convention and do not interpret its
direction here. Official MUSE evaluation was completed only for seed 42, so
these differences should not be interpreted as statistically stable.

\begin{table}[H]
\centering
\caption{MUSE-News seed-42 held-out diagnostics. MIA AUC is diagnostic; Band is
the fraction inside the calibrated target interval.}
\label{tab:muse-internal}
\footnotesize
\renewcommand{\arraystretch}{1.08}
\setlength{\tabcolsep}{3.8pt}
\begin{tabular}{@{}lcccc@{}}
\toprule
Method & \makecell{Forget\\NLL} & \makecell{Retain\\NLL$\downarrow$}
& \makecell{MIA\\AUC$\downarrow$} & \makecell{Band\\fraction$\uparrow$}\\
\midrule
Original & 0.576 & 0.761 & 0.981 & 0.015\\
GA       & 1.865 & 2.018 & \best{0.934} & 0.071\\
NPO      & 1.858 & 2.010 & \best{0.934} & 0.072\\
GradDiff & 1.379 & \best{1.546} & 0.961 & 0.081\\
SimNPO   & 1.513 & 1.597 & 0.955 & \best{0.144}\\
\method{}& 1.557 & 1.666 & 0.949 & 0.131\\
\bottomrule
\end{tabular}
\end{table}

\begin{table}[H]
\centering
\caption{Official MUSE-News seed-42 metrics. PrivLeak follows the evaluator's
sign convention and is not direction-ranked.}
\label{tab:muse-official}
\small
\renewcommand{\arraystretch}{1.08}
\setlength{\tabcolsep}{4.5pt}
\begin{tabular}{@{}lcccc@{}}
\toprule
& \multicolumn{2}{c}{Forget$\downarrow$} & Retain$\uparrow$ & \\
\cmidrule(lr){2-3}\cmidrule(lr){4-4}
Model & VerbMem & KnowMem & KnowMem & PrivLeak\\
\midrule
Original & 57.42 & 64.09 & 54.12 & -99.81\\
SimNPO   & \best{19.27} & \best{45.74} & 38.27 & -98.32\\
\method{}& 20.71 & 45.83 & \best{38.72} & -96.52\\
\bottomrule
\end{tabular}
\end{table}
\subsection{Target-Band Ablations}

\begin{figure}[t]
\centering
\begin{tikzpicture}
\begin{axis}[
  width=\columnwidth,
  height=0.76\columnwidth,
  ybar stacked,
  bar width=8pt,
  ymin=0, ymax=1,
  ylabel={Fraction of forget examples},
  symbolic x coords={2S-40,2S-50,1S-40,1S-50,R-50},
  xtick=data,
  xticklabel style={rotate=30,anchor=east,font=\scriptsize},
  yticklabel style={font=\scriptsize},
  label style={font=\small},
  legend style={
    font=\scriptsize,
    legend columns=3,
    at={(0.5,1.03)},
    anchor=south
  },
  grid=major,
  grid style={draw=gray!20}
]
\addplot[fill=warningorange!70,draw=warningorange] coordinates {
  (2S-40,.575) (2S-50,.280) (1S-40,.435) (1S-50,.125) (R-50,.905)
};
\addplot[fill=targetgreen!55,draw=targetgreen] coordinates {
  (2S-40,.125) (2S-50,.160) (1S-40,.095) (1S-50,.080) (R-50,.020)
};
\addplot[fill=deepblue!65,draw=deepblue] coordinates {
  (2S-40,.300) (2S-50,.560) (1S-40,.470) (1S-50,.795) (R-50,.075)
};
\legend{Below,Inside,Above}
\end{axis}
\end{tikzpicture}
\caption{MUSE validation-screen target occupancy. 2S and 1S denote two-sided
and lower-only objectives; the suffix is the checkpoint step, and +R denotes
sparse retain replay. The upper target reduces overshoot at step 40, while
fixed final checkpoints overshoot and retain replay under-forgets.}
\label{fig:occupancy}
\end{figure}
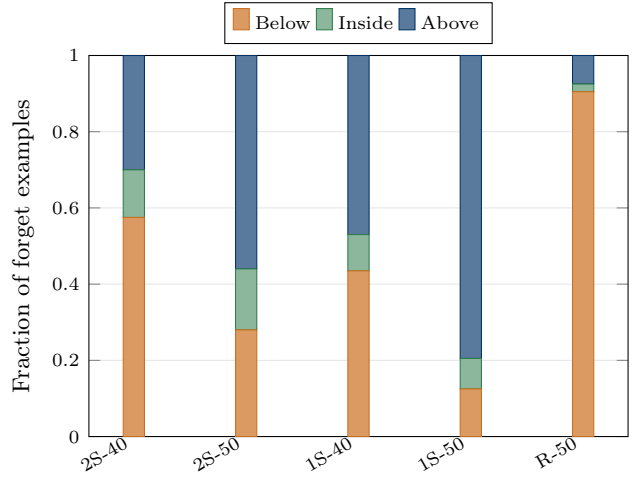

The upper target reduces overshoot (Table~\ref{tab:ablations} and
Fig.~\ref{fig:occupancy}): at step 40, the two-sided objective places 30.0\% of
examples above the band, compared with 47.0\% for the lower-only variant, and
improves retain NLL from 1.790 to 1.598. Continuing to the fixed final checkpoint
substantially increases the above-band fraction and retention damage for both
objectives. Sparse retain replay with weight 0.1 moves in the opposite direction
and leaves 90.5\% of examples below target. These results motivate target-aware
checkpoint selection rather than always using the final training step.

The corrected MUSE multiseed screen selects steps 40, 50, and 50 for seeds 42,
43, and 44, respectively. Their forget NLLs are 1.530, 0.956, and 1.531. Seed
43 leaves 93.0\% of examples below target, showing that a fixed 50-step budget
does not reliably reach the intended region.

\begin{table}[H]
\centering
\caption{MUSE validation-screen ablations. Occupancy columns are fractions
relative to the calibrated band.}
\label{tab:ablations}
\scriptsize
\renewcommand{\arraystretch}{1.08}
\setlength{\tabcolsep}{1.8pt}
\begin{tabular}{@{}lcccccc@{}}
\toprule
Variant & Step & \makecell{Forget\\NLL} & \makecell{Retain\\NLL}
& Below & Inside & Above\\
\midrule
Two-sided       & 40 & 1.530 & 1.598 & 0.575 & 0.125 & 0.300\\
Two-sided final & 50 & 1.800 & 1.852 & 0.280 & 0.160 & 0.560\\
Lower-only      & 40 & 1.711 & 1.790 & 0.435 & 0.095 & 0.470\\
Lower-only final& 50 & 2.107 & 2.171 & 0.125 & 0.080 & 0.795\\
Two-sided+R     & 50 & 1.107 & 1.195 & 0.905 & 0.020 & 0.075\\
\bottomrule
\end{tabular}
\rowcolors{2}{white}{white}
\end{table}

\section{Analysis and Discussion}

\paragraph{Computational cost.}
Let $\Fcal(b)$ denote the cost of a forward pass on batch size $b$, with backward
cost approximately $2\Fcal(b)$. GA, SimNPO, and \method{} each require one
current-model forget forward/backward pass, approximately $3\Fcal(|B_f|)$ per
step. NPO additionally requires stored original losses or a reference
computation, depending on implementation. GradDiff adds a retain forward/backward
pass every step. \method{} shifts its additional work to one-time target
calibration.

\paragraph{What the target band controls.}
The objective directly controls per-example loss relative to a non-member
reference distribution. It does not certify removal of training-data influence,
nor does matching losses guarantee matching all behavioral or privacy properties.
The official MUSE results illustrate this distinction: internal loss diagnostics
are useful for selection, but do not determine VerbMem, KnowMem, or PrivLeak.

This caution is consistent with theoretical and empirical work arguing that
unlearning should be judged not only by empirical forget loss, but also by
generalization, resource constraints, and stronger downstream evaluations
\citep{sekhari2021remember,pawelczyk2025machine}.

\paragraph{Calibration uncertainty.}
For $n$ calibration examples in a conditioning bin, the
Dvoretzky--Kiefer--Wolfowitz inequality gives, with probability at least
$1-\delta$,
\begin{equation}
  \sup_z|\widehat F_n(z)-F(z)|
  \leq\sqrt{\frac{\log(2/\delta)}{2n}}.
\end{equation}
This motivates coarse bins and minimum bin sizes. The bound addresses sampling
error, not distribution mismatch: a large but mismatched calibration set can
still define the wrong target.

\paragraph{Practical implication.}
The experiments suggest that the upper target and checkpoint rule are as
important as the lower forgetting target. A lower-only objective can still
overshoot, and retain replay can prevent target attainment if weighted too
strongly. A useful deployment protocol should therefore report target occupancy
through training and validate checkpoint selection on held-out data.

\section{Related Work}

\paragraph{Classical machine unlearning.}
Machine unlearning was originally motivated by the need to remove selected
training examples without retraining a model from scratch \citep{cao2015towards}.
Early work studied both exact and approximate deletion mechanisms. SISA training
reduces the cost of deletion by partitioning training into shards and slices
\citep{bourtoule2021machine}, while certified-removal methods provide stronger
guarantees in restricted model classes \citep{ginart2019making,guo2020certified,
neel2021descent}. Other approaches modify, reverse, or approximate training
updates to remove the influence of targeted examples
\citep{graves2021amnesiac,thudi2022unrolling,izoo2021approximate}. These works
establish the central tension between deletion quality, computational cost, and
retained utility, but they do not directly solve unlearning for modern generative
language models.
Sekhari et al. study machine unlearning from a generalization perspective,
emphasizing how many samples can be deleted while still maintaining performance
on unseen data under computation and storage constraints
\citep{sekhari2021remember}. This viewpoint is complementary to our empirical
LLM setting: H2DF controls a calibrated loss target during unlearning, but does
not provide the kind of distributional retraining equivalence or generalization
guarantee studied in theoretical unlearning.

\paragraph{Unlearning for large language models.}
LLM unlearning is harder because the forget target is often semantic,
generative, and difficult to localize. Gradient-ascent-style objectives are
simple and widely used, but they can cause utility collapse or abnormal
generations \citep{yao2023large,zhang2024negative}. Recent methods therefore add
retain losses, distillation, preference-style objectives, or optimization
constraints to better trade off forgetting and utility
\citep{chen2023unlearn,liu2024rethinking,wang2024rkld,pan2024multiobjective}.
Our method follows this line of work but focuses on a different control
mechanism: rather than maximizing forget loss indefinitely, it defines a
data-calibrated target region for each forget example.

\paragraph{Preference- and representation-based methods.}
Negative Preference Optimization reframes unlearning as a preference-style
objective and slows catastrophic collapse compared with GA
\citep{zhang2024negative}. SimNPO simplifies this formulation by removing the
reference-model dependence and obtains strong empirical results on TOFU and MUSE
\citep{fan2024simplicity}. Representation-based methods such as RMU steer
internal activations away from forget-domain representations and are especially
important for hazardous-knowledge removal \citep{li2024wmdp}. Follow-up work
shows that apparent forgetting can fail under stronger robustness and
knowledge-extraction tests \citep{lynch2024eight}. \method{} is
complementary: it does not directly steer hidden states or use preference pairs,
but instead calibrates the loss region that forgotten examples should reach.

\paragraph{Benchmarks and evaluation.}
TOFU provides a controlled fictitious-author QA benchmark for testing whether a
fine-tuned model behaves as if selected author profiles had not been learned
\citep{maini2024tofu}. MUSE evaluates unlearning across verbatim memorization,
knowledge memorization, privacy leakage, utility, scalability, and sustainability
\citep{shi2024muse}. WMDP evaluates hazardous-knowledge removal and introduced
representation misdirection as a strong unlearning baseline \citep{li2024wmdp}.
Other evaluations study membership inference, extraction, and memorization risks
in language models \citep{carlini2021extracting,carlini2023quantifying,
mattern2023membership}. These benchmarks show that high forget loss alone is not
sufficient: unlearning methods must also preserve retain behavior and avoid new
privacy or utility failures. This motivates \method{}'s calibrated target-band
view.

\section{Limitations}

The study has four main limitations. First, MUSE official metrics are available
only for seed 42, and the validation screens reveal substantial seed sensitivity.
Second, MUSE-Books, sequential requests, and larger-scale models are not
evaluated. Third, TOFU and MUSE use different base models and evaluation regimes,
so their numerical results are not directly comparable. Fourth, the loss-based
target is a proxy rather than a certified unlearning guarantee; performance
depends on the quality and representativeness of the calibration set. The
released RMU implementation uses LoRA rather than the original full-parameter
setup and is therefore excluded from the main quantitative comparison.

Recent robustness work further shows that practical unlearning methods may fail
under stronger evaluations such as data-poisoning removal, even when they appear
effective under standard metrics \citep{pawelczyk2025machine}. This supports our
choice to report H2DF as a controllable target-band mechanism rather than a
certified deletion method.

\section{Conclusion}

We introduced \method, a calibrated target-band formulation for language-model
unlearning. It preserves GA's early update direction while suppressing pressure
after examples reach a non-member-derived loss region. On TOFU, \method{} matches
the forgetting--retention tradeoff of GradDiff across three seeds with low
variability. On MUSE-News, it improves retention over GA and NPO and is
competitive with, but does not outperform, SimNPO. The upper target reduces
overshoot, while multiseed results show that checkpoint selection and
optimization stability remain open practical problems. Calibrated target bands
are therefore best viewed as a mechanism for controllable forgetting, not as a
guarantee of complete data removal.
\balance
\bibliographystyle{abbrvnat}
\bibliography{references}

\onecolumn
\appendix

\section{Training Algorithm}
\label{app:algorithm}

\begin{algorithm}[H]
\caption{Calibrated target-band unlearning. Calibration is completed before
optimization; targets are not re-estimated from evaluation data.}
\label{alg:h2df}
\begin{algorithmic}[1]
\Require Original model $\theta_0$; forget set $D_f$; matched calibration set
$D_{\mathrm{cal}}$; optional retain set $D_r$; quantiles
$\rho_{\mathrm{low}},\rho_{\mathrm{high}}$; $\tau,\gamma,\lambda$; replay
period $K$; training steps $T$
\Ensure LoRA adapter, fixed targets, diagnostics, and selected checkpoint
\State Score $D_f$ and $D_{\mathrm{cal}}$ with $\theta_0$; compute length,
original-loss, and available domain/type features
\State Build coarse feature bins; progressively back off conditioning features
when a matched bin is smaller than the minimum bin size
\For{each forget example $i$}
  \State Compute $a_i$ and optional $b_i$ from matched calibration quantiles
  \If{task is TOFU}
    \State $a_i\gets\max\{a_i,s_i(\theta_0)+\Delta_{\min}\}$
  \EndIf
\EndFor
\State Initialize trainable LoRA parameters $\theta$
\For{$t=1,\ldots,T$}
  \State Sample $B_f\subset D_f$ and compute per-example scores $s_i(\theta)$
  \State $\Lcal\gets\Lcal_{\mathrm{H2DF}}(\theta;B_f)$ from
  Eq.~\eqref{eq:h2df}
  \If{$D_r$ is used and $t\bmod K=0$}
    \State $\Lcal\gets\Lcal+\beta_r\Lcal_{\mathrm{retain}}(\theta;B_r)$
  \EndIf
  \State Backpropagate, clip gradients, update $\theta$, and log target occupancy
\EndFor
\State Select a saved checkpoint on the fixed validation screen; evaluate once
on held-out data
\end{algorithmic}
\end{algorithm}

\section{Reproducibility Details}
\label{app:reproducibility}

\begin{table}[H]
\centering
\caption{Primary configurations used for the reported experiments.}
\label{tab:appendix-config}
\small
\renewcommand{\arraystretch}{1.14}
\setlength{\tabcolsep}{6pt}
\rowcolors{2}{softgray}{white}
\begin{tabular}{@{}L{0.23\textwidth}L{0.34\textwidth}L{0.34\textwidth}@{}}
\toprule
\rowcolor{white}
Setting & TOFU & MUSE-News \\
\midrule
Base model/checkpoint
  & Qwen2.5-1.5B-Instruct with TOFU SFT adapter
  & MUSE-News target checkpoint \\
Tokenizer
  & model tokenizer
  & Llama-2-7B tokenizer \\
LoRA rank / alpha / dropout
  & $8/16/0$
  & $8/16/0$ \\
Target modules
  & \multicolumn{2}{c}{q, k, v, o, gate, up, and down projections} \\
Precision
  & BF16
  & BF16 \\
Batch / accumulation
  & $4/1$
  & $1/4$ \\
Maximum sequence length
  & 512
  & 1024 \\
Optimizer / schedule
  & AdamW / cosine
  & AdamW / cosine \\
Learning rate / steps
  & $10^{-4}/500$
  & $2\times10^{-5}/50$ \\
Warmup / gradient clipping
  & $25/1.0$
  & $25/1.0$ \\
$\tau$ / $\lambda$
  & $0.25/10^{-4}$
  & $0.25/10^{-4}$ \\
Calibration quantiles
  & $\rho=0.7$, $\Delta_{\min}=0.5$
  & $(\rho_{\mathrm{low}},\rho_{\mathrm{high}})=(0.5,0.9)$ \\
Upper-band weight $\gamma$
  & 0
  & 0.5 \\
Retain replay
  & every 10 steps for reported \method{} variant
  & none for the main \method{} run \\
\bottomrule
\end{tabular}
\rowcolors{2}{white}{white}
\end{table}

\begin{table}[H]
\centering
\caption{Evidence provenance. ``Screen'' denotes the fixed 200-example
checkpoint-diagnostic subset rather than final held-out evaluation; * indicates
availability for only a subset of methods.}
\label{tab:appendix-evidence}
\small
\renewcommand{\arraystretch}{1.16}
\setlength{\tabcolsep}{6pt}
\rowcolors{2}{softgray}{white}
\begin{tabular}{@{}L{0.22\textwidth}L{0.09\textwidth}L{0.31\textwidth}L{0.27\textwidth}@{}}
\toprule
\rowcolor{white}
Evidence block & Seeds & Evaluation scope & Manuscript use \\
\midrule
TOFU full QA
  & 42--44*
  & 200 forget; 3,420 retain examples
  & Tables~\ref{tab:tofu42}--\ref{tab:tofu3}, Fig.~\ref{fig:tofu-tradeoff} \\
MUSE internal comparison
  & 42
  & 888 forget; 1,778 retain; 1,521 calibration
  & Table~\ref{tab:muse-internal} \\
MUSE official suite
  & 42
  & Official VerbMem, KnowMem, and PrivLeak evaluation
  & Table~\ref{tab:muse-official} \\
MUSE multiseed screen
  & 42--44
  & Fixed 200-example validation screen
  & Seed-sensitivity discussion \\
MUSE ablation screen
  & 42
  & Fixed 200-example validation screen
  & Table~\ref{tab:ablations}, Fig.~\ref{fig:occupancy} \\
\bottomrule
\end{tabular}
\rowcolors{2}{white}{white}
\end{table}

\section{Checkpoint Selection and Reporting}
\label{app:selection}

For the short-budget MUSE runs, let
\begin{equation}
  r_t=\frac{1}{|V_f|}\sum_{i\in V_f}
  \mathbf{1}\!\left[s_i(\theta_t)<a_i\right]
\end{equation}
be the fraction of validation forget examples still below their lower target at
checkpoint $t$. We selected
\begin{equation}
  t^\star=\arg\min_{t\in\mathcal{T}} |r_t-0.5|,
\end{equation}
using a fixed 200-example validation screen and the saved checkpoint set
$\mathcal{T}$. This criterion was chosen before full held-out evaluation to
avoid selecting directly on the reported test metrics. It is a practical
heuristic rather than a theoretically optimal rule. The manuscript therefore
reports target occupancy alongside final benchmark metrics and distinguishes
validation-screen ablations from full held-out and official evaluations.

\section{Artifact Contents}
\label{app:artifacts}

The released artifact contains the training and calibration implementation,
benchmark-specific YAML configurations, baseline runners, CPU unit tests, and
the compact JSON metrics used in the manuscript. Model weights, adapters,
downloaded datasets, credentials, and raw experiment caches are intentionally
excluded. The expected workflow is:
\begin{enumerate}[label=\arabic*.,leftmargin=1.8em]
  \item run calibration to create fixed per-example targets;
  \item train \method{} or a configured baseline;
  \item evaluate the selected adapter on forget, retain, and calibration data;
  \item run the official MUSE evaluator for benchmark-level metrics.
\end{enumerate}

\end{document}